\documentclass[11pt]{article}

\usepackage[preprint]{acl}

\usepackage{times}
\usepackage{latexsym}
\usepackage[T1]{fontenc}
\usepackage[utf8]{inputenc}
\usepackage{microtype}
\usepackage{inconsolata}

\usepackage{graphicx}
\usepackage{subcaption}
\usepackage{booktabs}
\usepackage{tabularx}
\usepackage{float}
\usepackage{algorithm}
\usepackage{algorithmic}
\usepackage{amsmath}
\usepackage{amssymb}
\usepackage{mathtools}
\usepackage{amsthm}
\usepackage[capitalize,noabbrev]{cleveref}

\theoremstyle{plain}

\theoremstyle{definition}

\theoremstyle{remark}

\title{AQL-Lite: Enabling Efficient Knowledge Acquisition for Arabic LLMs via Data-Centric Continual Pretraining}

\author{
  \textbf{Houssam Eddine Boukhalfa}\textsuperscript{1}\thanks{\ \ Core contributor.}\quad
  \textbf{Mourad Bouache}\textsuperscript{2}\quad
  \textbf{Mourad Dahmane}\textsuperscript{2} \\
  \textsuperscript{1}National School of Artificial Intelligence (ENSIA), Algiers, Algeria \\
  \textsuperscript{2}Nojoom AI \\
  \small{\texttt{houssam-eddine.boukhalfa@ensia.edu.dz},\ \texttt{bouache@nojoom.ai},\ \texttt{mdahmane@gmail.com}}
}

\begin{document}
\maketitle

\begin{abstract}
Training Arabic large language models faces challenges from noisy web data, dialectal variation, and the high cost of continual pretraining (CPT). We introduce AQL-Lite (AQL: \emph{Arabic Quality Language}), a time- and resource-efficient framework for adapting a multilingual LLM to Modern Standard Arabic (MSA). AQL-Lite couples a quality-first curation pipeline (structural cleaning, heuristic and model-based filtering, MSA selection, and deduplication) with a partial-parameter CPT recipe that trains RS-LoRA adapters across the network together with the last eight transformer blocks and the embedding matrix, under layer-wise learning rates. At a matched 1B-token budget, CPT on AQL-curated data raises a four-benchmark Arabic mean by 3.51 points over CPT on raw web text, which yields no gain over the base model. In a controlled 5B-token comparison, AQL-Lite recovers 94\% of the gain of full-parameter CPT (mean 73.19 vs.\ 73.58) in 52\% of the wall-clock time and outperforms adapter-only baselines on every benchmark, and a component ablation shows that adapters, unfrozen blocks, and embedding training each contribute. Scaled to 40B MSA tokens and instruction-tuned on one million examples, the resulting 12B model scores 75.3 on the seven-benchmark Open Arabic LLM Leaderboard v2 (OALL v2) suite.
\end{abstract}

\section{Introduction}

Large language models (LLMs) have significantly advanced natural language processing, yet Arabic remains comparatively under-served despite its wide use and linguistic diversity. This gap is driven not only by model architecture, but also by two practical bottlenecks: \emph{data quality} and \emph{training cost}. Arabic web corpora contain large amounts of noisy, duplicated, templated, and low-signal text, while also reflecting variation across Modern Standard Arabic (MSA), dialects, code-switching, and inconsistent orthography. As a result, scaling raw Arabic tokens alone is insufficient for building strong Arabic LLMs.

Recent Arabic LLMs such as Jais and ALLaM show that strong Arabic performance requires careful preprocessing, filtering, deduplication, and distribution control \cite{jais2023,allam2024}. At the same time, continual pretraining (CPT) remains costly, especially when updating all model parameters. Parameter-efficient methods such as LoRA reduce this cost \cite{hu2021lora}, but adapter-only CPT may under-update the representations that a new language distribution requires \cite{biderman2024lora,pezeshkpour2025learning}. This motivates our central question: \emph{can we approximate the benefits of full Arabic CPT while substantially reducing training time and compute?}

We introduce \textbf{AQL-Lite} (AQL: \emph{Arabic Quality Language}), a data-centric framework for efficiently adapting a strong multilingual checkpoint to MSA. AQL-Lite first builds a high-quality corpus through a five-stage pipeline: structural cleaning, heuristic filtering, MSA/dialect control, model-based quality and safety scoring, and exact/near deduplication. It then performs \emph{partial-parameter} CPT: RS-LoRA adapters on all linear layers, full-rank updates to the last eight transformer blocks and the embedding matrix, and layer-wise learning rates. This design sits between adapter-only CPT, which is efficient but limited, and full-parameter CPT, which is effective but computationally expensive.

We evaluate AQL-Lite in three controlled settings. First, at a matched 1B-token budget, we vary only the training data (raw, heuristically filtered, or fully curated) to measure the effect of curation. Second, at 5B tokens, we compare the recipe against full CPT, LoRA-only, and RS-LoRA on identical data and hardware, and ablate its components. Third, we scale the recipe to 40B MSA tokens, instruction-tune the result, and evaluate the 12B model before and after instruction tuning on the seven-benchmark OALL v2 suite \cite{oall2_2025}.

\paragraph{Contributions.}
Our contributions are:
\begin{itemize}
  \item \textbf{Arabic quality-first data pipeline.}
  We introduce the AQL pipeline, a five-stage MSA corpus curation pipeline combining structural cleaning, heuristic filtering, MSA/dialect control, model-based quality and safety scoring, and exact/near deduplication, and show at a matched token budget that the curated data adds 3.51 points over raw web text.

  \item \textbf{Partial-parameter CPT.}
  We propose AQL-Lite, a CPT recipe that combines RS-LoRA, selective upper-block unfreezing, embedding training, and layer-wise learning rates to approximate full-CPT gains at lower cost, and ablate its components.

  \item \textbf{Empirical evaluation for MSA adaptation.}
  We compare AQL-Lite against full and adapter-only CPT in a controlled 5B-token setting, then evaluate the final 12B model on the OALL v2 suite, separating the contributions of CPT and instruction tuning.
\end{itemize}
\section{Related Work}

\paragraph{Arabic LLMs.}
Recent Arabic-centric LLMs show that strong Arabic capability is feasible, but also highlight persistent bottlenecks around data curation, dialect balance, and safety. Jais emphasizes Arabic preprocessing, deduplication, and bilingual training to mitigate Arabic data scarcity \cite{jais2023}. ALLaM reports strict language-ID filtering, spam/URL filtering, and exact deduplication, while noting challenges related to fuzzy deduplication, toxic content, and social bias in internet-scale corpora \cite{allam2024}. Broader surveys further document the rapid development of Arabic LLMs and the need for reliable Arabic evaluation practices \cite{mashaabi2024survey}.
Closest to our setting, AceGPT continually pretrains an English-centric base model on Arabic before localized instruction tuning \cite{huang2024acegpt}, and SambaLingo studies how to adapt English-centric LLMs to new languages, including Arabic, through continual pretraining with vocabulary extension \cite{csaki2024sambalingo}. Unlike these works, we adapt a multilingual base model and focus on reducing the cost of the continual-pretraining stage itself.

\paragraph{Data quality, filtering, and deduplication.}
A growing body of work shows that data quality substantially improves knowledge acquisition per token, often rivaling gains from model scaling. Large-scale filtering, repetition removal, and deduplication are now standard in strong pretraining pipelines \cite{rae2021scaling,penedo2023refinedweb}. These findings are especially relevant for Arabic, where raw web tokens are abundant but effective tokens are reduced by boilerplate, templated pages, SEO spam, duplication, orthographic variation, and dialect/code-switching effects.
For Arabic specifically, the 101 Billion Arabic Words dataset applies cleaning, normalization, and deduplication to Common Crawl \cite{aloui2024arabic101b}, and FineWeb2 adapts language identification, filtering, and deduplication to more than 1{,}000 languages, ablating each step with small-model training runs \cite{penedo2025fineweb2}. AQL adds explicit MSA/dialect control, a learned multi-category safety scorer, and per-document audit logs to this line of work.

\paragraph{Continual pretraining and parameter-efficient adaptation.}
Continual or domain-adaptive pretraining adapts a general checkpoint to a new distribution and can improve in-domain performance without training from scratch \cite{gururangan2020dontstop}. However, full-weight CPT is costly at LLM scale. Parameter-efficient fine-tuning methods such as LoRA reduce adaptation cost by learning low-rank updates \cite{hu2021lora}, but adapter-only CPT can under-update representations needed for robust knowledge acquisition. Recent work studies LoRA variants and their stability, including rank-stabilized scaling \cite{rslora2023}, retention and forgetting behavior \cite{biderman2024lora}, and broader PEFT trade-offs \cite{mao2025survey}. Most directly related, \citet{pezeshkpour2025learning} analyze how far LoRA-based continual pretraining can support domain-specific knowledge acquisition.
The individual ingredients of our recipe have precedents: LongLoRA shows that training the embedding and normalization layers alongside LoRA is important when adapting to a new context-length regime \cite{chen2024longlora}, and discriminative fine-tuning assigns different learning rates to different layers \cite{howard2018ulmfit}. Our contribution is to combine these with selective upper-block unfreezing for language-adaptive continual pretraining and to measure the resulting cost--quality trade-off against full CPT.

\paragraph{Arabic evaluation and safety benchmarks.}
Arabic LLMs are commonly evaluated using exam-style QA, multiple-choice knowledge benchmarks, and safety/trust stress tests. Representative benchmarks include AlGhafa \cite{alghafa2023}, ArabicMMLU \cite{arabicmmlu2024}, Arabic EXAMS \cite{exams2020}, and AraTrust \cite{aratrust2025}.
The Open Arabic LLM Leaderboard v2 combines these with MadinahQA, ALRAGE, and ArbMMLU-HT into a seven-benchmark suite \cite{oall2_2025}, and HELM Arabic re-implements the same suite within the HELM framework \cite{helm2022,helm_arabic_leaderboard}.

\section{Proposed Method} 
\subsection{AQL Data Pipeline}
\label{sec:aql_pipeline}

Arabic web-scale text is abundant but highly heterogeneous: boilerplate, templated pages, SEO spam, aggressive duplication, and inconsistent orthography reduce the fraction of \emph{effective} training tokens. AQL Pipeline is a quality-first pipeline that converts noisy Arabic web text into a curated corpus suitable for continual pretraining (CPT). We start from \textbf{885B} raw Arabic tokens and filter to a \textbf{215B}-token high-quality corpus. \Cref{tab:token_accounting} summarizes token retention.

\paragraph{Design goals.}
AQL is optimized for (i) \textbf{token efficiency} (maximize downstream gains per token), (ii) \textbf{distribution control} (explicitly manage MSA vs.\ dialect and code-switching), (iii) \textbf{auditability} (store per-stage reasons/scores), and (iv) \textbf{safety-by-construction} (filter risky content \emph{before} CPT).

\subsubsection{Data model and streaming execution}
\label{subsec:aql_datamodel}
AQL operates on sharded JSONL where each record contains:
\texttt{\{doc\_id, url, timestamp, source, raw\_text, meta\}}.
Each stage appends derived fields (e.g., \texttt{clean\_text}, \texttt{features}, \texttt{msa\_prob}, \texttt{quality\_score}, \texttt{safety\_scores}, \texttt{dedup\_cluster}) and a structured \texttt{filter\_log} capturing pass/fail reasons. The pipeline is implemented in a streaming manner to avoid materializing full corpora in memory: each stage reads shards, emits filtered shards, and writes lightweight counters for token accounting and ablations.

\paragraph{Stage 1: Structural Cleaning and Arabic Normalization.}
Stage 1 removes structural noise that inflates token counts without adding linguistic signal, producing \texttt{clean\_text}:
\textbf{Markup stripping:} remove HTML tags, script/style blocks, and embedded navigation fragments when detectable.
\textbf{Boilerplate cleanup:} drop repeated headers/footers, cookie banners, and templated lists using simple repetition heuristics.
\textbf{Unicode and whitespace:} normalize Unicode, remove control characters, collapse whitespace/newlines, and standardize punctuation spacing.
\textbf{Arabic orthography normalization:} remove tatweel, and normalize common punctuation variants.
This stage also extracts and preserves metadata (URL/domain, language-id hints if available) to support domain-level dedup and analysis.
\begin{figure*}[t]
  \centering
  \vspace{-2mm}
  \includegraphics[width=0.98\textwidth]{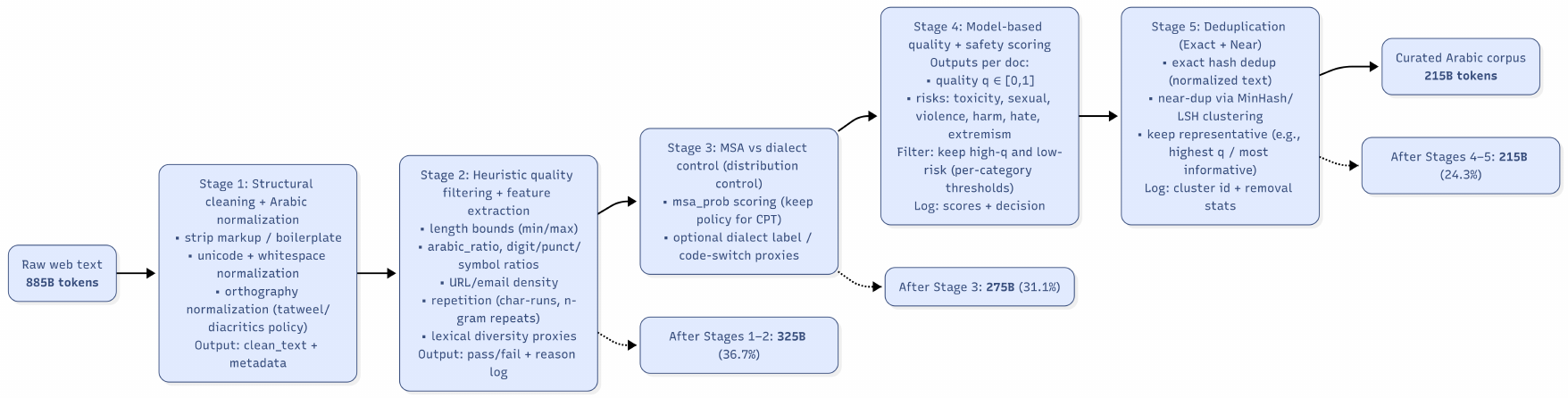}
  \vspace{-2mm}
  \caption{\textbf{AQL Pipeline overview.} The AQL pipeline progressively refines raw web text via structural cleaning, heuristic filtering, dialect control, model-based quality/safety scoring, and exact+near deduplication.}
  \label{fig:pipeline}
  \vspace{-2mm}
\end{figure*}
\begin{table}[t]
\centering
\small
\setlength{\tabcolsep}{4pt}
\begin{tabularx}{\columnwidth}{@{}Xrr@{}}
\toprule
Stage & Tokens (B) & Retention \\
\midrule
Raw crawl & 885 & 100.0\% \\
Stages 1--2 (structural + heuristic quality) & 325 & 36.7\% \\
Stage 3 (MSA selection) & 275 & 31.1\% \\
Stages 4--5 (model-based + dedup) & 215 & 24.3\% \\
\bottomrule
\end{tabularx}
\caption{Token accounting through the AQL pipeline.}
\label{tab:token_accounting}
\end{table}

\paragraph{Stage 2: Heuristic Quality Filtering with Feature Computation.}
Stage~2 computes lightweight document-level features on \texttt{clean\_text} (Stage~1 output) and removes clearly low-signal pages using \textbf{predefined thresholds} (decided \emph{before} training and logged for reproducibility), following the general principle that large-scale web corpora benefit substantially from systematic filtering and de-dup–aware quality controls \cite{penedo2024fineweb,rae2021scaling}. The intent is to eliminate common Arabic web failure modes that inflate token counts without adding learning signal, including spam/SEO artifacts, boilerplate-heavy templates, digit/symbol-heavy noise, and low-diversity repetitive text. Concretely, we enforce basic \textbf{length and layout constraints} (min/max tokens or characters, and mean line length), compute \textbf{script/noise ratios} (Arabic character ratio, digit ratio, punctuation ratio, and symbol ratio), and quantify \textbf{repetition and diversity} (maximum repeated-character runs, within-document repeated token $n$-gram fraction, and lexical diversity proxies such as unique-token ratio). We also include simple \textbf{lexical/template indicators}, such as stopword-ratio bounds, a template score based on near-identical lines or list-like patterns, and a \textbf{banned-keyword trigger} that flags documents containing terms from a small curated blacklist of high-risk/low-signal keywords; this blacklist is treated as a deterministic rule and is versioned and logged like other thresholds. Both the stopword list and the blacklist were built by the authors. For auditability, every rejected document is assigned a \textbf{primary rejection reason} (e.g., \texttt{low\_arabic\_ratio}, \texttt{repetition}, \texttt{template}, \texttt{banned\_keyword}), and we store the computed features and the final decision, enabling debugging and corpus analysis.

\begin{table}[t]
\centering
\small
\setlength{\tabcolsep}{4pt}
\begin{tabularx}{\columnwidth}{@{}Xr@{}}
\toprule
\textbf{Rule (on \texttt{clean\_text})} & \textbf{Threshold} \\
\midrule
Min / max tokens per document & $50\,/\,100{,}000$ \\
Arabic character ratio$^{\dagger}$ & $\geq 0.70$ \\
Digit + punct.\ + symbol ratio & $\leq 0.30$ \\
Max repeated-character run & $\leq 10$ \\
Repeated token 4-gram fraction & $\leq 0.40$ \\
Unique-token ratio & $\geq 0.30$ \\
Stopword ratio & $[0.15,\,0.45]$ \\
Banned keyword present & drop \\
\bottomrule
\end{tabularx}
\caption{Stage~2 heuristic rules. Thresholds were fixed before any training run. $^{\dagger}$This rule also removes technical and code-switched MSA with many Latin-script tokens (see Limitations).}
\label{tab:heuristics}
\end{table}

\paragraph{Stage 3: MSA vs.\ Dialect Classification.}
Arabic performance depends strongly on dialect balance and code-switching. Stage~3 assigns each document an MSA probability $p(\mathrm{MSA}\mid d)$ using a MARBERTv2 classifier \cite{abdulmageed2021arbertmarbert} fine-tuned on 70K manually annotated samples (\Cref{app:msa_dialect_full}) and applied to all Stage~2 survivors. We apply a \textbf{100\% MSA policy}: only documents accepted as MSA by the classifier are retained, which keeps 84.6\% of Stage~2 tokens (275B of 325B) and makes the curated 215B-token corpus MSA-only.

\paragraph{Stage 4: Model-Based Quality \& Safety Scoring.}
Stage~4 applies a learned multi-task scorer to each candidate document to estimate both \textbf{content quality} and \textbf{safety risk}. Concretely, we use a \textbf{Gemma~3 4B} \cite{team2025gemma} model fine-tuned on \textbf{100K labeled samples} to produce a 7-dimensional score vector for each document $d$:
\begin{multline*}
\mathrm{score}(d)=\big(q,\ s_{\text{tox}},\ s_{\text{sex}},\ s_{\text{viol}},\\
s_{\text{harm}},\ s_{\text{hate}},\ s_{\text{ext}}\big)\in[0,1]^7,
\end{multline*}
where $q$ is a continuous \emph{quality} score (higher is better), and $s_\cdot$ are category-wise \emph{risk scores} (higher indicates a higher likelihood of unsafe content in that category). We retain documents that are \textbf{high quality} while keeping \textbf{risk scores below thresholds}:
\[
\text{keep}(d)=\mathbb{1}\big[q \ge \tau_q \,\wedge\, \forall k\in\mathcal{K}:\ s_k \le \tau_s(k)\big],
\]
with a quality threshold $\tau_q$, per-category safety thresholds $\tau_s(k)$, and $\mathcal{K}=\{\text{tox},\text{sex},\text{viol},\text{harm},\text{hate},\text{ext}\}$. Threshold values are listed in \Cref{tab:stage4_thresholds}.

\paragraph{Stage 5: Exact + Near-Duplicate Removal.}
Duplication is pervasive in Arabic web data (mirrors, reposts, scraped news, and templated pages), which wastes CPT steps and reduces effective diversity. AQL therefore applies deduplication \emph{after} the strongest filtering stages so that similarity is computed over cleaner candidates and the dedup policy preserves informative variety. For \textbf{exact dedup}, we compute a deterministic hash over a normalized representation of \texttt{clean\_text} (collapsed whitespace and standardized punctuation) and remove duplicates using a persistent hash index. For \textbf{near-duplicate dedup}, we use a standard MinHash+LSH pipeline: documents are shingled into $n$-grams, mapped to MinHash signatures, and bucketed via locality-sensitive hashing to identify near-copies above a Jaccard threshold $\tau_{dup}$. Within each near-duplicate cluster, we keep a single representative according to a deterministic policy based on the Stage~4 quality score, content length, and timestamp, and drop the rest. We log cluster sizes, removal rates, and representative-selection statistics to quantify how much token reduction is attributable to duplication and to support auditability.

\paragraph{Final corpus format.}
The final corpus is stored as \textbf{shuffled, length-bucketed JSONL shards} for efficient training. Each record includes \texttt{text} plus minimal metadata (e.g., \texttt{doc\_id}, source/domain, \texttt{msa\_prob}, Stage~4 quality/safety scores, and dedup cluster ids) to support mixture control and reproducible token accounting. \Cref{alg:aql} summarizes the full pipeline.

\label{subsec:aql_pseudocode}
\begin{algorithm}[t]
\caption{AQL pipeline: a streaming filtering pass followed by global deduplication.}
\label{alg:aql}
\small
\begin{algorithmic}[1]
\STATE \textbf{Input:} raw shards $\{S_k\}$; \textbf{Output:} curated shards $\{S'_k\}$
\STATE $\mathcal{C} \leftarrow \emptyset$ \COMMENT{pool shared by all shards}
\FOR{each document $d$ in each shard $S_k$}
  \STATE $x \leftarrow \mathrm{Clean}(d.\mathrm{raw})$;\ \ $f \leftarrow \mathrm{Features}(x)$
  \IF{\textbf{not} $\mathrm{HeuristicPass}(f)$} \STATE \textbf{continue} \COMMENT{Stage 2, log reason} \ENDIF
  \IF{\textbf{not} $\mathrm{MSAPass}(p(\mathrm{MSA}\mid x))$} \STATE \textbf{continue} \COMMENT{Stage 3} \ENDIF
  \STATE $(q,s) \leftarrow \mathrm{Scorer}(x)$
  \IF{\textbf{not} $\mathrm{keep}(q,s)$} \STATE \textbf{continue} \COMMENT{Stage 4} \ENDIF
  \STATE add $d$ to $\mathcal{C}$ with $x, f, q, s$
\ENDFOR
\STATE $\mathcal{C} \leftarrow \mathrm{NearDedup}(\mathrm{ExactDedup}(\mathcal{C}))$ \COMMENT{Stage 5}
\STATE write $\mathcal{C}$ as shuffled shards $\{S'_k\}$ with audit metadata
\end{algorithmic}
\end{algorithm}

\subsection{AQL-Lite: Partial-Parameter Continual Pretraining}
\label{sec:aql_lite}

The AQL pipeline produces a high-quality MSA corpus, but continual pretraining (CPT) with full-weight updates remains expensive at 12B scale. AQL-Lite approximates full CPT with a \textbf{partial-parameter} recipe that updates low-rank adapters everywhere and full-rank weights only where they are expected to matter most.

\paragraph{What we did end-to-end.}
Our training follows four phases: \textbf{(1)} controlled CPT experiments that vary only the data at a matched \textbf{1B-token} budget and only the training recipe at a matched \textbf{5B-token} budget; \textbf{(2)} the main CPT scale-up on \textbf{40B tokens} sampled from the MSA-only AQL corpus; \textbf{(3)} supervised instruction tuning on \textbf{1M} examples (\textbf{800K} core instruction-following + \textbf{200K} safety/alignment); and \textbf{(4)} deployment with a lightweight \textbf{safety classifier} that screens unsafe user inputs before they reach the model. All large-scale training is executed on \textbf{8$\times$A100 80GB} GPUs.

\subsubsection{Training objective}
Let $\theta_0$ be a pretrained base checkpoint. CPT adapts $\theta_0$ to a target distribution $\mathcal{D}$ (here: AQL-curated MSA text) by minimizing next-token prediction loss:
\begin{equation}
\min_{\theta}\; \mathbb{E}_{x \sim \mathcal{D}}\Big[\mathcal{L}_{\text{NTP}}(x;\theta)\Big].
\label{eq:cpt_objective}
\end{equation}
Full CPT updates all parameters $\theta$, which is costly, while adapter-only CPT can under-update the representations a new language distribution requires. AQL-Lite balances the two by training LoRA adapters across the network and unfreezing only the last \textbf{8 transformer blocks}, together with the token embedding matrix.
Within the unfrozen blocks, layer-wise learning rates assign larger steps to later layers; the blocks below them are updated only through low-rank adapters. This lets high-level representations adapt aggressively while lower-level linguistic features change little.

\subsubsection{Partial-parameter CPT}
\label{subsec:structured_peft}

In a single CPT run, AQL-Lite jointly trains
\textbf{(i) RS-LoRA adapters} on all linear projections (query, key, value, output, gate, up, and down),
\textbf{(ii) all weights of the last 8 transformer blocks}, and
\textbf{(iii) the token embedding matrix}.
We regulate adaptation with \textbf{differentiated learning rates}: separate rates for adapters and embeddings, and a layer-wise schedule within the unfrozen blocks that gives larger updates to later blocks.

\paragraph{Trainable parameter partition.}
Let the base model parameters be $\theta$, adapter parameters be $\phi$, token embeddings be $E$, and the number of transformer blocks be $L$ ($L=48$ for Gemma~3 12B). We unfreeze the last $N=8$ blocks, with index set
\begin{equation}
\mathcal{U}=\{L-7,\,\dots,\,L\},
\end{equation}
and partition base parameters as
\begin{equation}
\theta = \theta_{\text{frozen}} \cup \theta_{\mathcal{U}},
\qquad \theta_{\mathcal{U}}=\{\theta_\ell:\ell\in\mathcal{U}\}.
\end{equation}
AQL-Lite trains the set
\begin{equation}
\Theta_{\text{train}}=\{\phi,\ \theta_{\mathcal{U}},\ E\}.
\end{equation}
The 8 unfrozen blocks contain 1.79B parameters and the embedding matrix, which Gemma~3 ties to the LM head and which we count once, 1.01B; together they are 2.80B, or 23.8\% of the 11.77B-parameter language model. RS-LoRA adapters on all seven projections of all 48 blocks add 4.09M parameters per unit of rank, so the total is $2.80\text{B} + r \times 4.09\text{M}$, where $r$ is the adapter rank. Even at a large rank such as $r=256$ the adapters add about 1.05B parameters, so the trainable share stays below one third of the model; \Cref{tab:hyperparams} lists the full-rank groups.

\paragraph{Objective.}
We optimize the standard next-token prediction (NTP) loss over sequences $x\sim\mathcal{D}$:
\begin{equation}
\min_{\phi,\theta_{\mathcal{U}},E}\ \ \mathbb{E}_{x\sim\mathcal{D}}
\Big[\mathcal{L}_{\text{NTP}}\big(x;\theta_{\text{frozen}},\theta_{\mathcal{U}},\phi,E\big)\Big].
\label{eq:aql_lite_objective}
\end{equation}

\paragraph{RS-LoRA adapters (stable low-rank updates).}
For a base linear projection $W\in\mathbb{R}^{d_{\text{out}}\times d_{\text{in}}}$, LoRA adds a low-rank update $W' = W + \Delta W$ with $\Delta W = \gamma_r\,AB$, where $A\in\mathbb{R}^{d_{\text{out}}\times r}$, $B\in\mathbb{R}^{r\times d_{\text{in}}}$, and $r\ll \min(d_{\text{out}},d_{\text{in}})$. Standard LoRA uses $\gamma_r=\alpha/r$; RS-LoRA \cite{rslora2023} uses
\begin{equation}
\gamma_r = \frac{\alpha}{\sqrt{r}},
\label{eq:lora}
\end{equation}
which keeps the scale of the update stable as the rank grows and improves CPT stability.

\paragraph{Selective unfreezing (fractional full-rank updates).}
Unlike full CPT (updates all $\theta$) or adapter-only CPT (updates only $\phi$), AQL-Lite additionally updates full-rank base weights in the upper blocks only:
\begin{equation}
\theta_\ell\ \text{is}\ \begin{cases}
\text{trainable}, & \ell\in\mathcal{U},\\
\text{frozen}, & \ell\notin\mathcal{U}.
\end{cases}
\label{eq:upper_unfreeze}
\end{equation}
This targets the blocks where high-level semantic abstractions are thought to form, while keeping lower-level lexical and syntactic processing stable.

\paragraph{Differentiated (layer-wise) learning rates.}
We use separate optimizer parameter groups for the adapters ($\eta_\phi$), the embeddings ($\eta_E$), and each unfrozen block $\ell\in\mathcal{U}$ ($\eta_\ell$). For the unfrozen blocks we use a monotone schedule that gives \emph{larger} steps to later layers, following discriminative fine-tuning \cite{howard2018ulmfit}:
\begin{equation}
\eta_\ell = \eta_{\text{top}}\cdot \gamma^{(L-\ell)},\qquad \ell\in\mathcal{U},\ \ \gamma\in(0,1).
\label{eq:layerwise_lr}
\end{equation}

\paragraph{Update rule.}
Let $\mathcal{G}\in\{\phi,\ E,\ \theta_\ell(\ell\in\mathcal{U})\}$ denote a parameter group with learning rate $\eta_{\mathcal{G}}$. Each group is updated by the optimizer with its own rate; schematically,
\begin{equation}
\mathcal{G}\leftarrow \mathcal{G} - \eta_{\mathcal{G}}\ \Delta_{\mathcal{G}},
\label{eq:param_group_update}
\end{equation}
where $\Delta_{\mathcal{G}}$ is the optimizer's update direction for $\mathcal{G}$.

\paragraph{Embedding training.}
We unfreeze the embedding matrix $E$ and train it with its own learning rate $\eta_E$. This adapts the representations of existing subwords to MSA text; the tokenizer is not extended. In our runs $\eta_E=3\times10^{-4}$, larger than $\eta_{\text{top}}=10^{-4}$.

\paragraph{Design rationale.}
Full-rank updates to the upper blocks and embeddings allow larger representational changes than low-rank adapters alone, while adapters in the frozen blocks provide a cheap path for the rest of the network to adjust. \Cref{subsec:ablations} tests each component.

\section{Experiments}
\subsection{Continual Pretraining}
We run controlled CPT experiments and one scale-up, all with the recipe of \Cref{sec:aql_lite}:
\begin{itemize}
  \item \textbf{Experiment 1 (5B tokens, recipe comparison):} AQL-Lite against full CPT and adapter-only CPT on the same \textbf{5B-token} subset, testing whether the recipe recovers most of the full-CPT gain at lower cost.
  \item \textbf{Ablations (\Cref{subsec:ablations}):} a data ablation at a matched \textbf{1B-token} budget, varying only the training data, and a component ablation at \textbf{5B tokens}, varying only the recipe.
  \item \textbf{Experiment 2 (40B tokens, scale-up):} the same recipe on \textbf{40B tokens} of the MSA-only AQL corpus, followed by instruction tuning.
\end{itemize}

\begin{figure*}[t]
  \centering
  \vspace{-2mm}
  \includegraphics[width=0.98\textwidth]{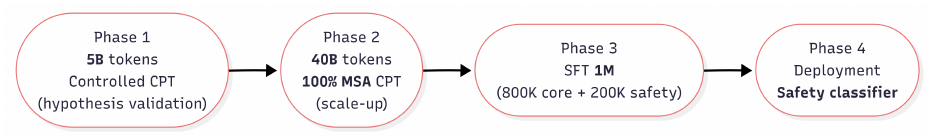}
  \vspace{-2mm}
  \caption{\textbf{AQL-Lite phases.} We validate the recipe in a 5B-token controlled CPT setting, then scale CPT to 40B tokens (100\% MSA), followed by 1M-example instruction tuning and deployment-time safety filtering.}
  \label{fig:lite}
  \vspace{-2mm}
\end{figure*}

\subsubsection{Experiment 1: 5B-token controlled CPT}
\label{subsec:exp1_5b}

\paragraph{Objective.}
We test whether partial-parameter CPT can recover most of the gains of full CPT while substantially reducing compute. To isolate the effect of the training recipe, all methods are trained on the \textbf{same fixed 5B-token subset} and evaluated with the same protocol.

\paragraph{Data and setup.}
The 5B-token subset is 96\% MSA text sampled from the AQL corpus, plus 2\% code and 2\% math data drawn from outside the AQL corpus. All runs use identical normalization and tokenization, identical hardware (\textbf{8$\times$A100 80GB}), and the same configuration: sequences packed into 4096-token blocks, gradient accumulation 16, bfloat16 precision, gradient clipping 1.0, and DeepSpeed ZeRO-2 \cite{rasley2020deepspeed}. \Cref{tab:hyperparams} summarizes the configuration.

\paragraph{Methods.}
We compare five settings: \textbf{Gemma 3 12B PT} \cite{team2025gemma} with no CPT, \textbf{Full CPT} with all parameters trainable, \textbf{LoRA-only} with adapter-only updates, \textbf{RS-LoRA} with rank-stabilized adapter-only updates, and \textbf{AQL-Lite}. AQL-Lite combines RS-LoRA on all linear layers, full-rank updates to the last eight blocks (2.80B parameters together with the embeddings), embedding training, and layer-wise learning rates with $\eta_{\phi}{=}2\times10^{-4}$, $\eta_{\text{top}}{=}10^{-4}$, $\gamma{=}0.7$, and $\eta_{E}{=}3\times10^{-4}$.

\paragraph{Evaluation.}
We evaluate all checkpoints on four Arabic benchmarks: \textbf{ArabicMMLU} \cite{arabicmmlu2024}, \textbf{AlGhafa} \cite{alghafa2023}, \textbf{AraTrust} \cite{aratrust2025}, and \textbf{Arabic EXAMS} \cite{exams2020}, and report the macro-average as Mean. All checkpoints in Tables~\ref{tab:exp1_results}--\ref{tab:component_ablation} are evaluated with the same protocol; these scores come from a separate evaluation of CPT checkpoints and should not be compared directly with \Cref{tab:helm_comparison}.

\begin{table*}[t]
\centering
\small
\setlength{\tabcolsep}{7pt}
\begin{tabular}{lcccccc}
\toprule
\textbf{Method} & \textbf{Time} & \textbf{ArabicMMLU} & \textbf{AlGhafa} & \textbf{AraTrust} & \textbf{Arabic EXAMS} & \textbf{Mean} \\
\midrule
Gemma 3 12B PT & --  & 65.56 & 60.68 & 87.01 & 54.38 & 66.91 \\
Full CPT       & 121h & \textbf{68.92} & 77.33 & \textbf{89.22} & \textbf{58.85} & \textbf{73.58} \\
LoRA-only      & 17h  & 66.43 & 71.11 & 87.75 & 55.68 & 70.24 \\
RS-LoRA        & 24h  & 66.57 & 72.81 & 88.27 & 55.87 & 70.88 \\
\textbf{AQL-Lite (ours)} & 63h & 68.26 & \textbf{78.00} & 89.15 & 57.36 & 73.19 \\
\bottomrule
\end{tabular}
\caption{Controlled CPT experiment on 5B tokens. All methods use the same data, hardware, and evaluation protocol. Scores are in percentage points; Mean is the macro-average across the four benchmarks. Time is wall-clock time on 8$\times$A100 80GB.}
\label{tab:exp1_results}
\end{table*}

\begin{figure}[t]
  \centering
  \includegraphics[width=\columnwidth]{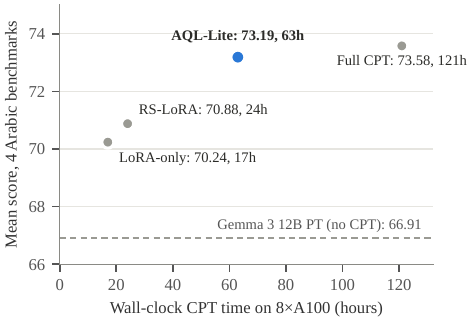}
  \caption{Mean score on four Arabic benchmarks against wall-clock CPT time for the 5B-token runs of \Cref{tab:exp1_results}.}
  \label{fig:cost_quality}
\end{figure}

\paragraph{Results and analysis.}
Table~\ref{tab:exp1_results} shows that AQL-Lite reaches a mean of \textbf{73.19}, compared with \textbf{73.58} for Full CPT and 66.91 for the base model. It thus recovers 94\% of the full-CPT gain (6.28 of 6.67 points) while reducing wall-clock training time from \textbf{121h} to \textbf{63h} (52\%). It outperforms LoRA-only and RS-LoRA on all four benchmarks. AQL-Lite is ahead of Full CPT on AlGhafa (78.00 vs.\ 77.33) and behind it on AraTrust (89.15 vs.\ 89.22); both margins are small, and we do not read them as differences between the methods. All rows are single training runs. \Cref{fig:cost_quality} plots score against training time.

Wall-clock time alone does not isolate where the savings come from: in every method, adapters in all blocks (and, for AQL-Lite, the trainable embeddings) require gradients to flow through the whole network, so the savings come from fewer weight gradients and smaller optimizer state rather than from a shorter backward pass.

\subsubsection{Ablations}
\label{subsec:ablations}

\begin{table}[t]
\centering
\footnotesize
\setlength{\tabcolsep}{2.5pt}
\resizebox{\columnwidth}{!}{%
\begin{tabular}{@{}lccccc@{}}
\toprule
\textbf{Data (1B tok.)} & \textbf{ArMMLU} & \textbf{AlGhafa} & \textbf{AraTr.} & \textbf{EXAMS} & \textbf{Mean} \\
\midrule
Base (no CPT) & 65.56 & 60.68 & 87.01 & 54.38 & 66.91 \\
Raw crawl & 65.37 & 59.32 & 88.06 & \textbf{54.46} & 66.80 \\
Stages 1--2 & 65.61 & 62.77 & 88.12 & 53.28 & 67.45 \\
Stages 1--5 (AQL) & \textbf{66.39} & \textbf{69.84} & \textbf{88.83} & 56.17 & \textbf{70.31} \\
\bottomrule
\end{tabular}}
\caption{Data ablation: CPT on 1B tokens of each data condition with an identical training recipe; only the data changes. ArMMLU = ArabicMMLU, AraTr.\ = AraTrust, EXAMS = Arabic EXAMS.}
\label{tab:data_ablation}
\end{table}

\begin{figure}[t]
  \centering
  \includegraphics[width=\columnwidth]{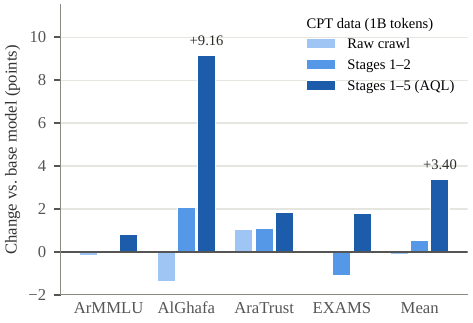}
  \caption{Data ablation: change over the base model after CPT on 1B tokens of each data condition (values from \Cref{tab:data_ablation}).}
  \label{fig:data_ablation}
\end{figure}

\paragraph{Does curation matter?}
\Cref{tab:data_ablation} varies only the training data at a matched 1B-token budget. CPT on raw crawled Arabic leaves the mean unchanged (66.80 vs.\ 66.91 for the base model). Structural cleaning and heuristic filtering (Stages 1--2) add 0.65 points, and the full pipeline adds a further 2.86, for a total of \textbf{+3.51} over raw text at the same budget (\Cref{fig:data_ablation}). Most of the gain comes from AlGhafa (+10.52 over raw), with smaller gains on Arabic EXAMS (+1.71), ArabicMMLU (+1.02), and AraTrust (+0.77). The mean increases at each stage, but not every benchmark does: Arabic EXAMS drops slightly after Stages 1--2. Because the last row applies MSA selection, model-based filtering, and deduplication together, this experiment does not separate their contributions.

\begin{table}[t]
\centering
\footnotesize
\setlength{\tabcolsep}{2.5pt}
\resizebox{\columnwidth}{!}{%
\begin{tabular}{@{}lccccc@{}}
\toprule
\textbf{Config.\ (5B tok.)} & \textbf{ArMMLU} & \textbf{AlGhafa} & \textbf{AraTr.} & \textbf{EXAMS} & \textbf{Mean} \\
\midrule
RS-LoRA only & 66.57 & 72.81 & 88.27 & 55.87 & 70.88 \\
Unfreeze last 8 only & 64.74 & 63.84 & 88.42 & 54.85 & 67.96 \\
RS-LoRA + emb. & 67.28 & 75.23 & 88.12 & 55.43 & 71.52 \\
RS-LoRA + unfreeze & 66.89 & 74.91 & 88.84 & 56.26 & 71.73 \\
AQL-Lite (full) & \textbf{68.26} & \textbf{78.00} & \textbf{89.15} & \textbf{57.36} & \textbf{73.19} \\
\midrule
\textit{Full CPT} & \textit{68.92} & \textit{77.33} & \textit{89.22} & \textit{58.85} & \textit{73.58} \\
\bottomrule
\end{tabular}}
\caption{Component ablation on the 5B-token subset of \Cref{tab:exp1_results}. ``Unfreeze'' trains all weights of the last eight blocks; ``emb.'' trains the embedding matrix. Full CPT is shown for reference.}
\label{tab:component_ablation}
\end{table}

\paragraph{Which components matter?}
\Cref{tab:component_ablation} adds the components of AQL-Lite one at a time. Unfreezing the last eight blocks without adapters is the weakest configuration (67.96), 2.92 points below RS-LoRA alone, so the network-wide low-rank path matters more than full-rank updates confined to the top blocks. Adding embedding training (+0.64) or the unfrozen upper blocks (+0.85) to RS-LoRA each helps, and the full recipe gains 2.31 points over RS-LoRA, more than the sum of the two individual gains (1.49). The full recipe also adds layer-wise learning rates, which are not ablated separately, so the remaining 0.82 points may reflect an interaction between components, the learning-rate schedule, or both. We also did not tune the number of unfrozen blocks; a sweep over $N\in\{4,8,12\}$ is left to future work.

\subsubsection{Experiment 2: 40B-token CPT scale-up (100\% MSA)}
\label{subsec:exp2_40b}

\paragraph{Objective and data policy.}
Experiment~2 applies the same recipe to \textbf{40B tokens} sampled from the MSA-only AQL corpus. The 40B budget is chosen to adapt an already capable multilingual checkpoint, not for compute-optimal pretraining, and we make no claim about the optimal token budget. The MSA-only setting reduces dialectal variance and targets the formal Arabic emphasized by our benchmark suite; MSA--dialect mixtures are left to future work.

\paragraph{Training configuration.}
We keep the configuration of Experiment~1 (\Cref{tab:hyperparams}) and scale the token budget. The resulting CPT checkpoint is then instruction-tuned to obtain \textbf{AQL-Lite 12B (instruct)}.

\subsection{Alignment}
\label{subsec:alignment}

We convert the 40B-token CPT checkpoint into the final assistant model using supervised instruction tuning (SFT) on \textbf{1M} Arabic instruction examples: \textbf{800K} core instruction-following examples and \textbf{200K} safety/alignment examples, mixed 80/20. The loss is cross-entropy computed only on assistant response tokens, following standard instruction-tuning practice \cite{chung2024scaling}. The data covers factual QA, summarization, extraction, rewriting, translation, classification, constrained generation, structured outputs, and multi-turn dialogue.

\paragraph{Data provenance.}
The data was built from Nojoom AI's production Arabic assistant: the prompts are human-written questions drawn from its usage logs, and the responses were crafted on the same platform.

\paragraph{Inference-time safety classifier.}
At deployment time, we add a lightweight safety classifier as an inference-time guardrail. It screens unsafe user requests before they reach AQL-Lite 12B (instruct), enabling refusal or safe redirection without retraining the LLM.

\section{Evaluation}
\label{sec:eval}

\subsection{Benchmarks and protocol}
\label{subsec:auto_eval}

We evaluate on the seven benchmarks of the Open Arabic LLM Leaderboard v2 (OALL v2) \cite{oall2_2025}: \textbf{AlGhafa} \cite{alghafa2023}, \textbf{ArabicMMLU} \cite{arabicmmlu2024}, \textbf{Arabic EXAMS} \cite{exams2020}, \textbf{MadinahQA} \cite{arabicmmlu2024}, \textbf{AraTrust} \cite{aratrust2025}, \textbf{ALRAGE} \cite{oall2_2025}, and \textbf{ArbMMLU-HT} \cite{jais2023}. We evaluate our own checkpoints with the OALL v2 evaluation suite. For reference, \Cref{tab:helm_comparison} also lists scores of other models on the same seven benchmarks as reported by the HELM Arabic leaderboard \cite{helm_arabic_leaderboard}; we did not run these models. HELM Arabic uses its own protocol (zero-shot generation with Arabic option letters, at most 1{,}000 instances per subset), so comparisons between our rows and the leaderboard rows are indicative rather than exact.

\begin{table*}[t]
\centering
\footnotesize
\setlength{\tabcolsep}{3pt}
\resizebox{\textwidth}{!}{%
\begin{tabular}{@{}lcccccccc@{}}
\toprule
\textbf{Model} & \textbf{Mean} & \textbf{AlGhafa} & \textbf{ArabicMMLU} & \textbf{Arabic EXAMS} & \textbf{MadinahQA} & \textbf{AraTrust} & \textbf{ALRAGE} & \textbf{ArbMMLU-HT} \\
\midrule
\multicolumn{9}{@{}l}{\textit{Reported by the HELM Arabic leaderboard}} \\
Gemini 2.5 Flash & \textbf{81.7} & 78.0 & \textbf{81.6} & \textbf{63.2} & \textbf{85.0} & 93.9 & 89.5 & \textbf{80.6} \\
GPT-5.1 (2025-11-13) & 80.9 & \textbf{79.1} & 80.3 & 63.1 & 80.4 & \textbf{94.1} & \textbf{90.9} & 78.4 \\
Qwen3 235B A22B Instruct & 78.6 & 77.3 & 76.9 & 59.0 & 77.7 & 91.7 & 89.0 & 78.3 \\
Llama 4 Scout Instruct & 74.6 & 77.5 & 73.4 & 60.0 & 72.5 & 89.1 & 81.1 & 68.6 \\
ALLaM-7B-Instruct-preview & 69.9 & 70.0 & 69.4 & 55.0 & 68.7 & 83.7 & 85.2 & 57.1 \\
Jais-adapted-70B-chat & 66.3 & 72.9 & 63.9 & 53.1 & 60.5 & 85.4 & 71.3 & 57.1 \\
\midrule
\multicolumn{9}{@{}l}{\textit{Gemma 3 12B base (evaluated by us with OALL v2)}} \\
AQL-Lite-12B (CPT only, before SFT) & 74.8 & 77.5 & 74.6 & 58.8 & 75.2 & 89.2 & 74.1 & 73.9 \\
\textbf{AQL-Lite-12B (after SFT)} & 75.3 & 78.2 & 75.0 & 59.2 & 75.7 & 89.7 & 74.8 & 74.3 \\
\bottomrule
\end{tabular}}
\caption{Head-to-head comparison on the seven OALL v2 benchmarks, in percentage points (higher is better; Mean is the macro-average). The bottom block shares the Gemma 3 12B base and was evaluated by us with OALL v2. The top block is taken from the HELM Arabic leaderboard, which uses a different protocol, so cross-block comparisons are indicative only. Best value per column in bold.}
\label{tab:helm_comparison}
\end{table*}

\subsection{Results}
\label{subsec:results}

With the protocol caveat above, AQL-Lite-12B's mean (75.3) lies between those reported for Qwen3 235B (78.6) and Llama 4 Scout (74.6), and above ALLaM-7B (69.9) and Jais-adapted-70B (66.3). It is close to the best reported AlGhafa score (78.2 vs.\ 79.1 for GPT-5.1) and comparatively weakest on ALRAGE (74.8), the only open-ended, passage-grounded task in the suite.

\paragraph{CPT versus instruction tuning.}
AQL-Lite-12B already scores 74.8 before SFT; instruction tuning adds 0.5 points to the mean and between 0.4 and 0.7 points on each benchmark. On this suite, the model's performance is therefore largely in place after CPT, and SFT mainly contributes instruction-following and safety behaviour that these benchmarks do not measure directly.

\section{Discussion}
\label{sec:discussion}

Our results suggest that efficient MSA adaptation depends on the interaction between \textbf{corpus quality}, \textbf{distribution control}, and \textbf{where model updates are applied}. AQL-Lite couples a quality-first data pipeline with a CPT recipe that updates both low-rank adapters and selected full-rank parameters.

\paragraph{Curation, not corpus size.}
The pipeline keeps 24.3\% of raw tokens, but the size of that reduction does not by itself show that the removed text was less useful. The controlled comparison does: at a matched 1B-token budget, raw web text leaves the base model unchanged while curated text adds 3.51 points (\Cref{tab:data_ablation}). For the MSA-centred benchmarks we use, effective token quality matters more than raw token volume.

\paragraph{Partial-parameter CPT as a practical middle ground.}
Adapter-only CPT is efficient but limited, while full CPT is effective but costly. AQL-Lite recovers 94\% of the full-CPT gain in 52\% of the wall-clock time. The component ablation indicates that the network-wide low-rank path is necessary, since full-rank updates confined to the top blocks perform worst, and that full-rank upper-block and embedding updates add to it.

\paragraph{Knowledge acquisition or re-alignment?}
Gemma~3 is a multilingual model that already handles Arabic, and our benchmarks cannot separate genuinely new knowledge from better use of knowledge the base model already has. The data ablation shows that curated text contributes more than raw text at the same budget, but it does not resolve this question. A from-scratch study on a smaller model (1B--4B parameters) would test the corpus outside the adaptation setting.

\paragraph{Safety as a layered design.}
AQL-Lite treats safety as a pipeline-level property rather than a single alignment step. Risk is reduced upstream through model-based corpus filtering, during SFT through a dedicated safety/alignment subset, and at inference time through a lightweight safety classifier. This layered design is especially relevant for Arabic deployments, where cultural, religious, regional, and political sensitivities can make unsafe or inappropriate outputs more harmful. We report AraTrust for the model alone and do not measure the contribution of each layer separately.

\paragraph{Future work.}
AQL-Lite-12B is the first model in the AQL family. Future work will study controlled MSA--dialect mixtures selected with the Stage~3 classifier, sweep the number of unfrozen blocks, run a from-scratch study on a smaller model, scale token budgets, improve morphology-aware filtering, and extend evaluation to dialect robustness, bias analysis, and deployment-oriented safety auditing.

\section*{Limitations}

AQL-Lite remains an efficiency--performance trade-off rather than a full replacement for full-weight continual pretraining: in the controlled 5B-token experiment it recovers 94\% of the full-CPT gain but does not match Full CPT on every benchmark. The main 40B-token scale-up uses a \textbf{100\% MSA} policy, which strengthens standardized Arabic performance but underrepresents dialectal Arabic, code-switching, and region-specific language use; the model should not be assumed to handle dialects well.

Several choices are not fully ablated. The number of unfrozen blocks ($N=8$) was not tuned, the layer-wise learning-rate schedule is not ablated separately from the other components, and the data ablation applies Stages 3--5 jointly. All results come from single training runs, so differences below about one point should not be over-interpreted. We have not yet measured $n$-gram overlap between our training data and the benchmarks, or the effect of CPT on English capabilities; both are planned for a future version. We also did not run full CPT at 40B tokens, so the efficiency comparison with full CPT is validated only at the 5B-token scale. Our efficiency comparison counts CPT wall-clock time only; it excludes the cost of running the Stage~3 and Stage~4 classifiers over the corpus and the cost of instruction tuning. As discussed in \Cref{sec:discussion}, our benchmarks also cannot distinguish new knowledge from re-alignment of knowledge the base model already has.

Our evaluation is based on automatic benchmarks and a head-to-head comparison in which the comparison rows are taken from a public leaderboard that uses a different evaluation protocol, rather than re-run by us. The MSA classifier's 99.96\% accuracy is measured in-domain; such a high value may partly reflect annotation artifacts or near-duplicates shared between the training and validation splits, and an out-of-domain check on dialect identification data such as MADAR \cite{bouamor2018madar} or NADI \cite{abdulmageed2020nadi} is needed to confirm it. The Stage~2 requirement of at least 70\% Arabic characters likely removes technical and code-switched MSA, such as programming tutorials and scientific text with Latin-script terms, which may limit the model on technical and coding prompts. The evaluation may also not capture all dimensions of Arabic instruction-following, dialect robustness, factual reliability, or deployment safety. Finally, despite corpus filtering, safety instruction tuning, and an inference-time safety classifier, the model may still hallucinate, reflect biases from web data, or fail under adversarial prompting, especially in high-stakes domains.

\section*{Ethics Statement}

AQL-Lite is intended to support Arabic language technology for education, information access, writing assistance, and Arabic-centric retrieval-augmented applications. Because Arabic NLP systems operate in culturally, religiously, and politically sensitive contexts, model errors or biased outputs may cause harm. Relevant risks include misinformation, cultural or religious insensitivity, dialect marginalization caused by MSA-centred training, and unsafe responses to adversarial prompts.

\paragraph{Instruction data.}
The instruction-tuning data was built from Nojoom AI's production Arabic assistant: the prompts come from its usage logs and the responses were crafted on the same platform. Because the prompts were written by real users, their use for training depends on user consent and on removing personal information, so this version does not release the instruction-tuning data.

\paragraph{Mitigations.}
We mitigate these risks through layered safeguards: upstream corpus filtering with model-based quality and safety scoring, safety-focused instruction tuning, and a deployment-time safety classifier that screens user inputs before they reach the model. However, these safeguards do not eliminate all risks. We therefore recommend that AQL-Lite not be used as the sole decision-maker in high-stakes settings such as medical, legal, financial, or civic decision-making. Any release should include clear usage guidelines, staged access when appropriate, monitoring for misuse, and continued auditing for bias, toxicity, dialect coverage, and culturally sensitive failure cases.

\bibliography{aql_lite_refs}

\clearpage
\appendix

\section{Pipeline Details}
\label{sec:appendix}

\subsection{Stage 3: MSA vs.\ Dialect Classifier}
\label{app:msa_dialect_full}

Stage~3 enforces distribution control by predicting whether a document is Modern Standard Arabic (MSA) or dialectal. This enables the \textbf{100\% MSA} policy used here and, in future work, controlled MSA/dialect mixtures.

\paragraph{Model and training.}
We fine-tune a MARBERTv2 encoder \cite{abdulmageed2021arbertmarbert} with a classification head on \textbf{70K} labeled samples.

\paragraph{Inference policy.}
Given a document $d$, the classifier outputs $p(\mathrm{MSA}\mid d)$, and documents accepted as MSA are kept. This retains 84.6\% of Stage~2 tokens (275B of 325B).

\paragraph{Validation performance.}
\Cref{tab:app_stage3_metrics_full} reports in-domain validation metrics. They measure agreement with our annotations on held-out samples from the same distribution, not general MSA/dialect identification ability. Because near-perfect in-domain scores can also arise from annotation artifacts or near-duplicate leakage across splits, we treat them as an upper bound; an out-of-domain evaluation on MADAR \cite{bouamor2018madar} or NADI \cite{abdulmageed2020nadi} is left to future work.

\begin{table}[h]
\centering
\small
\setlength{\tabcolsep}{6pt}
\begin{tabular}{lc}
\toprule
\textbf{Metric} & \textbf{Value} \\
\midrule
Eval loss & 0.00346 \\
Accuracy & 0.99964 \\
Macro-F1 & 0.99963 \\
Macro-Precision & 0.99956 \\
Macro-Recall & 0.99970 \\
\bottomrule
\end{tabular}
\caption{MSA vs.\ dialect classifier, in-domain validation metrics (final epoch).}
\label{tab:app_stage3_metrics_full}
\end{table}

\subsection{Stage 4: Quality \& Safety Scorer}
\label{app:safety_scorer}

Stage~4 uses a learned multi-task scorer that outputs a continuous \emph{quality} score and category-wise \emph{safety risk} scores. We fine-tune a \textbf{Gemma~3 4B} backbone with a lightweight multi-head prediction layer: one regression head for quality and six classification heads for safety categories (toxicity, sexual, violence, harm, hate, extremism). Writing $\hat{q}$ for the predicted quality and $s_k$ for the predicted risk in category $k$, we optimize a weighted multi-task objective:
\begin{equation}
\mathcal{L} = \lambda_q \cdot \mathrm{MSE}(q,\hat{q}) + \sum_{k \in \mathcal{K}} \lambda_k \cdot \mathrm{BCE}(y_k, s_k),
\end{equation}
where $y_k$ is the binary label for category $k$ and the weights $\lambda$ are tuned to mitigate class imbalance and stabilize training.

\paragraph{Validation performance.}
\Cref{tab:app_stage4_metrics} reports held-out validation metrics.

\begin{table}[h]
\centering
\small
\setlength{\tabcolsep}{4pt}
\begin{tabular}{lcc}
\toprule
\textbf{Metric} & \textbf{Value} & \textbf{Notes} \\
\midrule
Eval loss & 0.0834 & Multi-task objective \\
Quality MSE & 0.00523 & Lower is better \\
Toxicity Acc / F1 & 0.958 / 0.958 & Binary head \\
Sexual Acc / F1 & 0.994 / 0.975 & Binary head \\
Violence Acc / F1 & 0.984 / 0.969 & Binary head \\
Harm Acc / F1 & 0.940 / 0.857 & Most ambiguous \\
Hate Acc / F1 & 0.976 / 0.939 & Binary head \\
Extremism Acc / F1 & 0.988 / 0.956 & Binary head \\
Safety Micro-F1 & 0.973 & All safety labels \\
\bottomrule
\end{tabular}
\caption{Stage~4 scorer validation metrics (final epoch). Quality is evaluated with MSE; safety categories report accuracy and F1.}
\label{tab:app_stage4_metrics}
\end{table}

\paragraph{Qualitative sanity checks.}
We run short, controlled checks to ensure that clearly unsafe inputs yield high risk scores while unrelated categories remain low. To avoid reproducing harmful text, we report aggregate metrics only.

\paragraph{Score thresholds.}
Stage~4 keeps a document if $q \ge \tau_q$ and $s_k \le \tau_s(k)$ for every category $k$. \Cref{tab:stage4_thresholds} lists the thresholds.

\begin{table}[h]
\centering
\small
\setlength{\tabcolsep}{6pt}
\begin{tabular}{lcc}
\toprule
\textbf{Rule} & \textbf{Condition} & \textbf{Threshold} \\
\midrule
Quality & $q \ge \tau_q$ & 0.7 \\
Toxicity & $s_{\text{tox}} \le \tau_s(\text{tox})$ & 0.5 \\
Hate & $s_{\text{hate}} \le \tau_s(\text{hate})$ & 0.5 \\
Harm & $s_{\text{harm}} \le \tau_s(\text{harm})$ & 0.4 \\
Sexual & $s_{\text{sex}} \le \tau_s(\text{sex})$ & 0.5 \\
Violence & $s_{\text{viol}} \le \tau_s(\text{viol})$ & 0.5 \\
Extremism & $s_{\text{ext}} \le \tau_s(\text{ext})$ & 0.5 \\
\bottomrule
\end{tabular}
\caption{Stage~4 filtering thresholds.}
\label{tab:stage4_thresholds}
\end{table}

\subsection{Stage 5: Deduplication}
\label{app:dedup}

Stage~5 reduces memorization risk and improves effective data diversity with \textbf{two-level deduplication}: (i) exact deduplication by hashing normalized text, and (ii) near-duplicate removal with shingling and MinHash/LSH.

\paragraph{Exact deduplication.}
We compute a stable hash over the Stage~1-normalized text and remove repeated items using a global index.

\paragraph{Near-duplicate deduplication.}
We represent each document as a set of shingles, approximate Jaccard similarity with MinHash signatures, and use LSH buckets to find candidate duplicates. Pairs with similarity above $\tau_{dup}$ are merged, keeping one representative per cluster; see \Cref{sec:aql_pipeline}.

\section{Training Configuration}
\label{app:hyperparams}

\begin{table}[h]
\centering
\small
\setlength{\tabcolsep}{4pt}
\begin{tabularx}{\columnwidth}{@{}Xr@{}}
\toprule
\textbf{Setting} & \textbf{Value} \\
\midrule
\multicolumn{2}{@{}l}{\textit{Shared by all 5B-token runs}} \\
Sequence length & 4096 (packed) \\
Precision & bfloat16 \\
Gradient clipping & 1.0 \\
Gradient accumulation & 16 \\
Parallelism & DeepSpeed ZeRO-2 \\
Hardware & 8$\times$A100 80GB \\
\midrule
\multicolumn{2}{@{}l}{\textit{AQL-Lite}} \\
Adapters & RS-LoRA, all linear layers \\
Unfrozen blocks & last 8 of 48 (1.79B) \\
Embeddings (tied LM head) & trained (1.01B) \\
$\eta_\phi$ (adapters) & $2\times10^{-4}$ \\
$\eta_{\text{top}}$ (top block) & $10^{-4}$ \\
$\gamma$ (layer-wise decay) & 0.7 \\
$\eta_E$ (embeddings) & $3\times10^{-4}$ \\
\midrule
\multicolumn{2}{@{}l}{\textit{Wall-clock time, 5B tokens}} \\
Full CPT / LoRA-only & 121h / 17h \\
RS-LoRA / AQL-Lite & 24h / 63h \\
\bottomrule
\end{tabularx}
\caption{Training configuration of the 5B-token runs in \Cref{tab:exp1_results}. The 40B-token run uses the AQL-Lite settings.}
\label{tab:hyperparams}
\end{table}

\section{Human Evaluation Figures}
\label{app:figures}

\begin{figure}[h]
  \centering
  \includegraphics[width=\columnwidth]{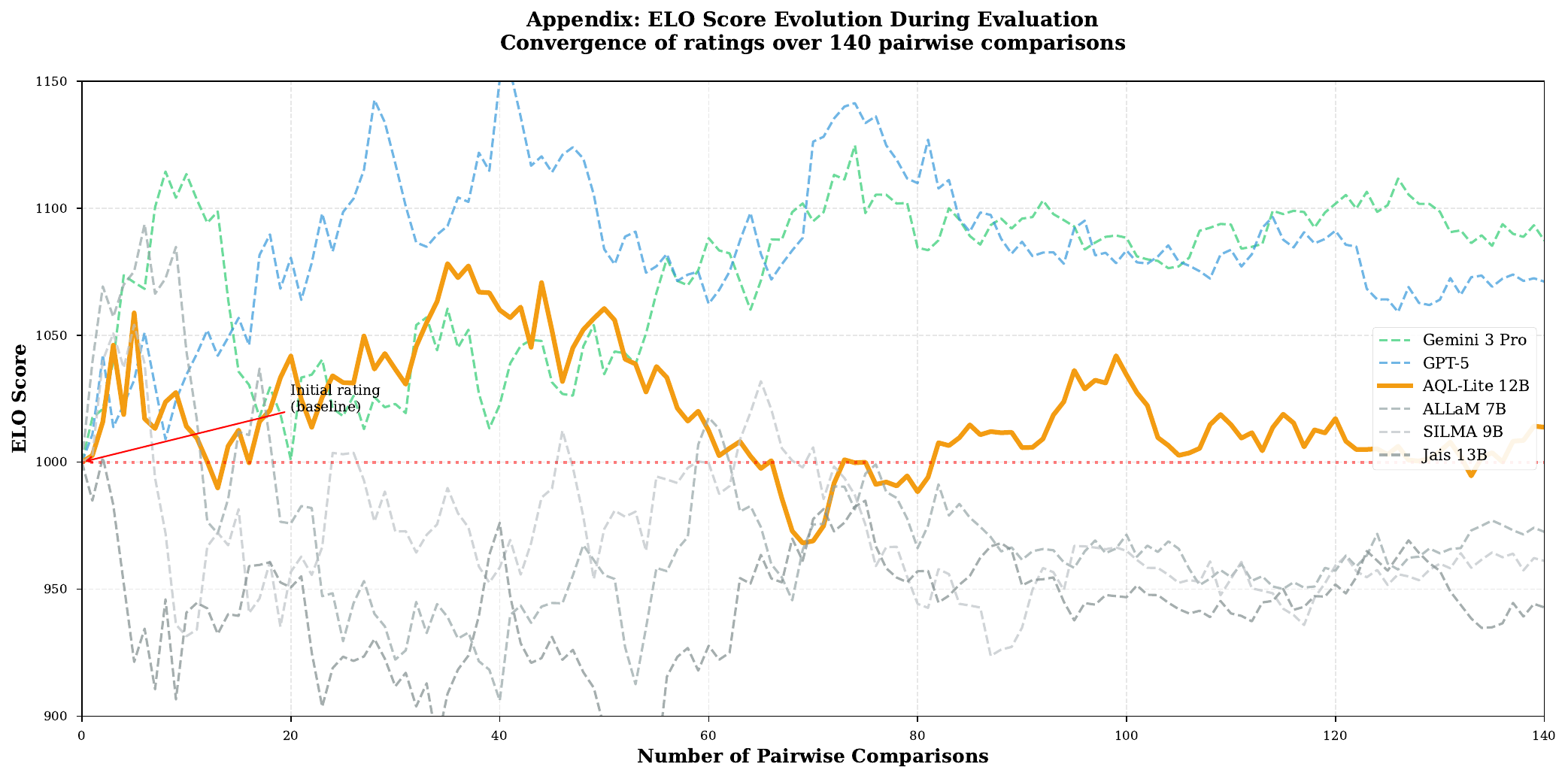}
  \caption{Evolution of Elo ratings over the course of the pairwise comparisons.}
  \label{fig:elo_evolution}
\end{figure}

\begin{figure*}[p]
  \centering
  \includegraphics[width=\textwidth]{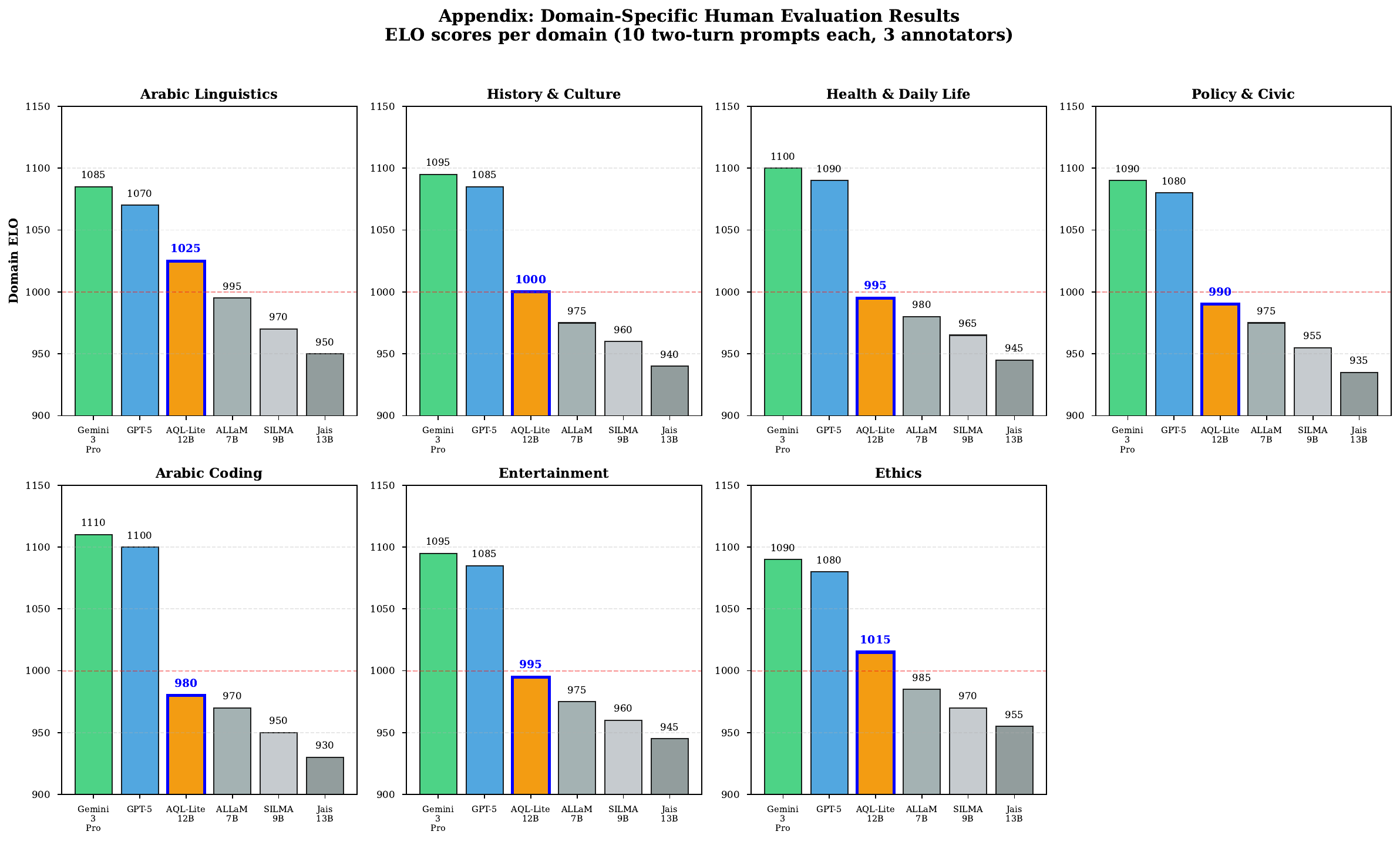}
  \caption{Per-domain Elo scores in the human evaluation (10 two-turn prompts per domain, 3 annotators).}
  \label{fig:domain_elo}
  \vspace{6mm}
  \includegraphics[width=0.92\textwidth]{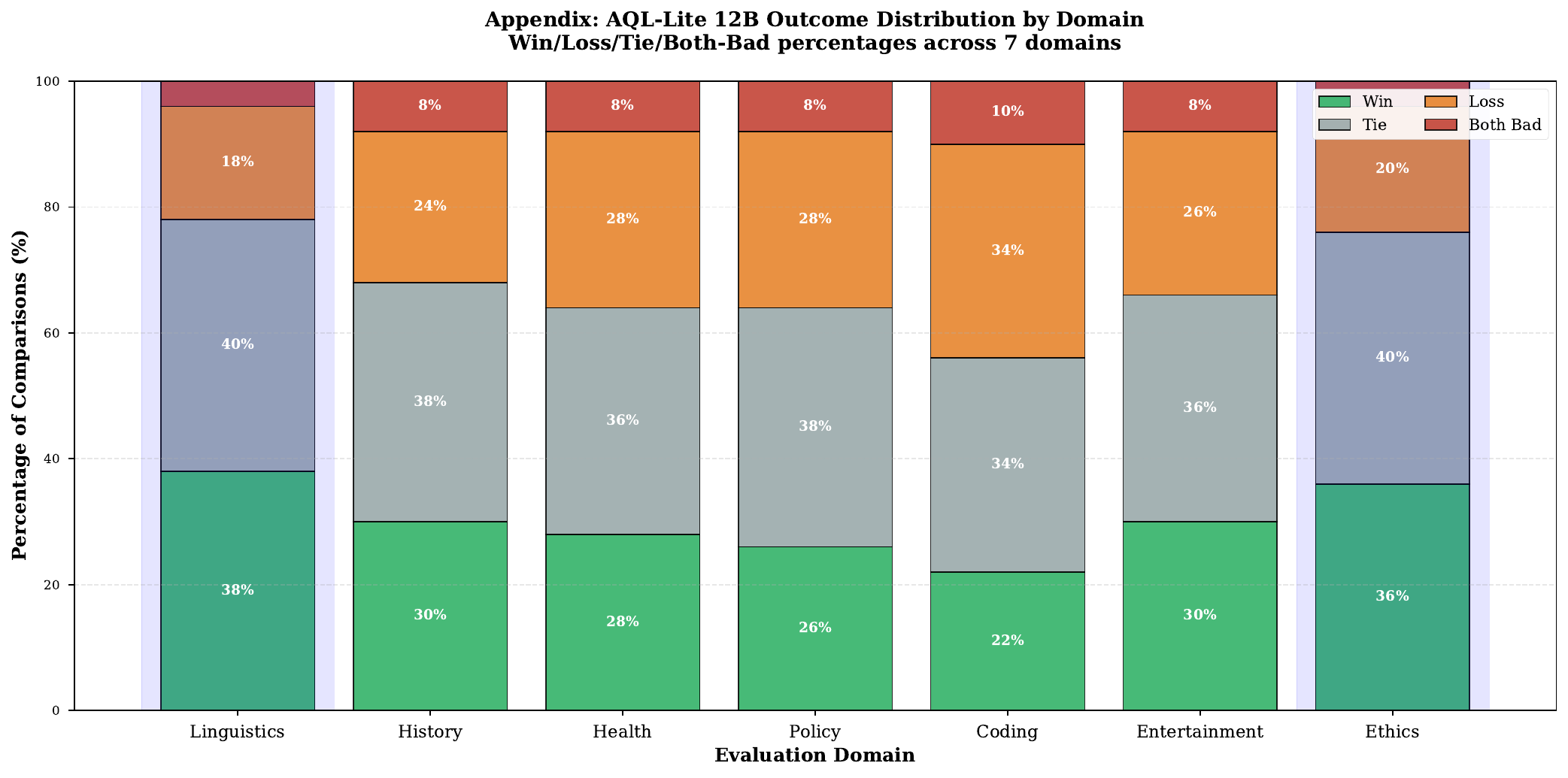}
  \caption{Outcomes of AQL-Lite-12B's pairwise comparisons (win, tie, loss, both bad) in each domain.}
  \label{fig:outcomes}
\end{figure*}

\Cref{fig:domain_elo,fig:outcomes,fig:elo_evolution} show per-domain results of the pairwise human evaluation.

\end{document}